\documentclass[10pt, onecolumn]{article}

\usepackage[utf8]{inputenc}
\usepackage{amsmath}
\usepackage{array}
\usepackage{graphicx}
\usepackage[hyphens]{url}
\usepackage{multirow}
\usepackage{hyperref}
\usepackage{float}
\usepackage{rotating}
\usepackage{bm}
\usepackage{xcolor}
\usepackage{subfig}
\usepackage{multicol}
\usepackage{authblk}

\usepackage{algorithm} 
\floatname{algorithm}{Algorithm}

\usepackage{algcompatible}

\algnewcommand\INPUT{\item[{\textbf{Input:}}]}
\algnewcommand\OUTPUT{\item[{\textbf{Output:}}]}
\algblockdefx{MAPP}{ENDMAPP}[1]%
  {\textbf{map partitions} #1}%
  {\textbf{end map}}
  
\algblockdefx{MAP}{ENDMAP}[1]%
  {\textbf{map} #1}%
  {\textbf{end map}}

\begin{document}
%
\title{An Information Theoretic Feature Selection Framework for Big Data under Apache Spark}

\author[1]{Sergio Ram\'irez-Gallego}
\author[2]{H\'ector~Mouri\~no-Tal\'in}
\author[2]{David~Mart\'inez-Rego}
\author[2]{Ver\'onica~Bol\'on-Canedo}
\author[1]{Jos\'e Manuel Ben\'itez}
\author[2]{Amparo~Alonso-Betanzos}
\author[1]{Francisco~Herrera}
\affil[1]{Department of Computer Science and Artificial Intelligence, University of Granada, 18071 Granada, Spain.
Emails: sramirez@decsai.ugr.es, j.m.benitez@decsai.ugr.es, herrera@decsai.ugr.es}
\affil[2]{Department of Computer Science, University of A Coru\~na, 15071 A Coru\~na, Spain.\\ Emails: h.mtalin@udc.es, dmartinez@udc.es, veronica.bolon@udc.es, ciamparo@udc.es}

\providecommand{\keywords}[1]{\textbf{\textit{Index terms---}} #1}

\maketitle

\begin{abstract}
With the advent of extremely high dimensional datasets, dimensionality reduction techniques are becoming mandatory. Among many techniques, feature selection has been growing in interest as an important tool to identify relevant features on huge datasets --both in number of instances and features--. The purpose of this work is to demonstrate that standard feature selection methods can be parallelized in Big Data platforms like Apache Spark, boosting both performance and accuracy. We thus propose a distributed implementation of a generic feature selection framework which includes a wide group of well-known Information Theoretic methods. Experimental results on a wide set of real-world datasets show that our distributed framework is capable of dealing with ultra-high dimensional datasets as well as those with a huge number of samples in a short period of time, outperforming the sequential version in all the cases studied.
\end{abstract}


%

%
%
%
%
\section{Introduction}
\label{sec:intro}

During the last few decades, the dimensionality of datasets employed in Machine Learning (ML) or Data Mining tasks has increased significantly. This presents an unprecedented challenge for researchers in these areas, since the existing algorithms not always respond in an adequate time when dealing with this new extremely high dimension. In fact, if we analyze the datasets posted in the popular libSVM Database~\cite{chang11}, we can observe that in the 1990s, the maximum dimensionality of the data was about 62\,000; in the 2000s, this number increased to more than 16 million; and in the 2010s it further increased to more than 29 million. In this new scenario, it is common now to deal with millions of features, so the existing learning methods need to be adapted.

With the advent of extremely high dimensional datasets mentioned above, the identification of the relevant features has become paramount. Dimensionality reduction techniques can be applied to reduce the dimensionality of the original data and even to improve learning performance \cite{guyon06, zhao2011spectral, bolon15b}. These dimensionality reduction techniques usually come in two flavors: \textit{feature selection} (FS) and \textit{feature extraction}. Both of them have their own merits. Feature extraction techniques combine the original features to yield a new set of features whereas feature selection techniques remove the irrelevant and redundant features. Due to the fact that FS maintains the original features, it is especially useful for applications where the original features are important for model interpretation and knowledge extraction~\cite{qiu13, bolon14}, and so this model will be the focus of this paper.

On the other hand, existing FS methods are not expected to scale well when dealing with Big Data due to the fact that their efficiency may significantly deteriorate or even become inapplicable~\cite{bolon15}. Scalable distributed programming protocols and frameworks have been appearing in the last decade to manage the problem of Big Data. The first programming model was MapReduce~\cite{dean04} along with its open-source implementation Apache Hadoop~\cite{whi12,had15}. Recently, Apache Spark~\cite{hamstra15, spa15}, a new distributed framework, was presented as a fast and general engine for large-scale data processing, popular among machine learning researchers due to its suitability for iterative procedures.

Likewise, several libraries for approaching ML task in Big Data environments have appeared in recent years. The first such library was Mahout~\cite{mah15}, subsequently followed by MLlib~\cite{mll15} built on top of the Spark system~\cite{spa15}. Thanks to Spark's ability to do in-memory computation and so speed up iterative processes, algorithms developed for this kind of platform become pervasive in industry. Despite the fact that several golden standard algorithms for ML tasks have been redesigned with a distributed implementation for big data technologies already, it is not the case for FS algorithms yet. Only a simple approach based on Chi-Squared\footnote{\url{http://spark.apache.org/docs/latest/mllib-feature-extraction.html}}, and an improvement to FS on Random Forest~\cite{sun14} have been proposed in the literature to deal with this problem.

This work aims at filling this gap. Our main purpose is to demonstrate that standard FS methods can be designed in these Big Data platforms and still can prove to be useful when dealing with big datasets, boosting both performance and accuracy. Here, we propose a new distributed design for a FS generic framework based on Information Theory~\cite{brown12}, which has been implemented using Apache Spark paradigm. A wide variety of techniques from the distributed environment have been used to make feasible this adaptation: information caching, data partitioning and replication of relevant variables, among others. Notice that adapting this framework to Spark implies a deep restructuring of these classic algorithms, which presents a big challenge for researchers. 

Lastly, to test the effectiveness of our framework, we have applied it to a complete set of real-world datasets (up to $O(10^7)$ features and instances). The subsequent results have shown the competitive performance (in terms of generalization performance and efficiency) of our method when dealing with huge datasets --both in number of features and instances--. As an illustrative example, we have been able to select 100 features in a dataset with $29 \times 10^6$ features and $19 \times 10^6$ instances in less than 50 minutes (using a 432-core cluster).

The remainder of this paper is organized as follows: Section 2 provides some background information about FS, Big Data, MapReduce programming model and other frameworks. Section 3 describes the distributed framework proposed for FS in Big Data. Section 4 presents and discuss the experiments carried out. Finally, Section 5 concludes the paper.

\section{Background}
\label{sec:backg}

In this section we give a brief introduction to FS, followed by a discussion on the advent of Big Data and its implication in this area. Finally, we outline the particularities of the MapReduce framework and the models derived from it.

\subsection{Feature Selection}

FS is a dimensionality reduction technique that tries to remove irrelevant and redundant features from the original data. Its goal is to obtain a subset of features that describes properly the given problem with a minimum degradation of performance, in order to obtain simpler and more accurate schemes~\cite{guyon06}.

Formally, we can define feature selection as: let $\bm{e_i}$ be an instance $\bm{e_i} = (e_{i1}, \ldots, e_{in}, e_{iy})$, where $e_{ir}$ corresponds to the $r$-th feature value of the $i$-th sample and $e_{iy}$ with the value of the output class $Y$. Let us assume a training set $D$ with $m$ examples, which instances $\bm{e_i}$ are formed by a set $X$ of $n$ characteristics or features, and a test set $D_t$ exist. Then let us define $S_\theta \subseteq X$ as a subset of selected features yielded by an FS algorithm. 

FS methods can be broadly categorized as~\cite{blum97}:

\begin{enumerate}
\item \textbf{Wrapper methods}, which use an evaluation function dependent on a learning algorithm~\cite{kohavi97}. They are aimed at optimizing a predictor as part of the learning process.
\item \textbf{Filtering methods}, which use other selection techniques as separability measures or statistical dependences. They only consider the general characteristics of the dataset, being independent of any predictor~\cite{guyon03}.
\item \textbf{Embedded methods}, which use a search procedure which is implicit in the classifier/regressor~\cite{saeys07}.
\end{enumerate}

Filter methods usually result in a better generalization due to its learning independence. Nevertheless, they usually select larger feature subsets, requiring sometimes a threshold to control them. Regarding complexity, filters are normally less expensive than wrappers. In those cases in which the number of features is large (especially for Big Data), it is indispensable to employ filtering methods as they are much faster than the other approaches. 

\subsection{Big Data: a two-sided coin}

Whereas the Internet continues generating quintillions bytes of data, the problem of handling large collections of those data is becoming more and more latent in our world.
For example, in 2012, 2.5 exabytes of daily data were created. Collecting, transmitting and maintaining large data is no longer feasible. 
Exceptional technologies are needed to efficiently process these large quantities of data to obtain information, within tolerable elapsed times.

Extracting valuable information from these collections of data has thus became one of the most important and complex challenges in data analytics research. This situation has caused that many knowledge extraction algorithms turn into obsolete methods when they face such vast amounts of data. As a result, the need for new methods, capable of managing such amount of data efficiently with similar performance, arises. 


Big Data is a popular term used to describe the exponential growth and availability of data nowadays, that becomes a problem for classical data analytics. Gartner~\cite{bey01} introduced the 3Vs concept by defining Big Data as high volume, velocity and variety information that require a new large-scale processing. Afterwards, this list was extended with 2 additional Vs. 
%
An under-explored but not less important topic is the ``Big Dimensionality'' in Big Data~\cite{zhai14}. This phenomenon, also known as the ``Curse of Big Dimensionality'', is boosted by the explosion of features and the combinatorial effects from new large incoming data where thousand or even millions of features are present. 

From the beginning, data scientists have generally focused on only one side of Big Data, which early days refers to the huge number of instances; paying less attention to the feature side. Big Dimensionality, though, calls for new FS strategies and methods that are able to deal with the feature explosion problem. It has been captured in many of the most famous dataset repositories in computational intelligence (like UCI or libSVM)~\cite{bache13,chang11}, where the majority of the new added datasets present a huge dimensionality~\cite{zhai14} (e.g. almost 30 millions for the dataset \emph{KDD2010}).

Not only the amount of features but also the myriad of feature types and their combinations are becoming a standard in many real world applications. For instance, on the Internet, all multimedia content represents about 60\% of total traffic~\cite{roush07}, transmitted in thousands of different formats (audio, video, images, etc.). Another example is Natural Language Processing (NLP), where multiple feature types such as words, $n$-gram templates, etc., are simultaneously employed so as to produce comprehensible and reliable models~\cite{mao13}. 

Despite this myriad, not all the features in a problem contribute equally on the prediction models and results. FS is thus required by the learning and prediction processes for a fast and cost-effective performance, now more than ever. Isolating high value features from the raw set of features (potentially irrelevant, redundant and noisy), while maintaining the requirements in measurement and storage, is one of the most important tasks in Big Data research.

\subsection{MapReduce Programming Model and Frameworks: Hadoop and Spark}
\label{subsec:mapreduce}

The MapReduce framework~\cite{dean04} was born in 2003 as a revolutionary tool in Big Data, designed by Google for processing and generating large-scale datasets. It was thought to automatically process data in a extremely distributed way through large clusters of computers. The framework is in charge of partitioning and managing data, recovering failure, job scheduling and communication; leaving to the programmers a transparent and scalable tool to easily execute tasks on distributed  systems\footnote{For a exhaustive review of MapReduce and others programming frameworks, please check~\cite{fern14}.}.

MapReduce is based on two phases of processing: Map and Reduce. First of all, users implement a Map function that processes key-value pairs which are transformed into a set of intermediate pairs, and a Reduce function that merges all intermediate pairs with a matching key. In this phase, the master node splits the data into chunks and distributes them across the nodes for an independent processing (in a divide-and-conquer fashion). Each node then executes the Map function on a given input subset and notifies its ending to the master node. After that, in the Reduce phase, the master node distributes the matching pairs across the nodes according to a key partitioning scheme, which combines these pairs using the Reduce function to form the final output.

The Map function takes $<$key, value$>$ pairs as input and yields a list of intermediate $<$key, value$>$ pairs as output. The Map function that internally process the data is defined by the user following a key-value scheme. Hence, the general scheme for a Map function is defined as: 

\begin{equation}\label{eq:map}
Map(<key1, val1>) \rightarrow list(<key2, val2>)
\end{equation} 

In the second phase, the master groups pairs by key and distributes the combined result to the Reduce functions started in each node. Here, a reduction function is applied to each associated list value and a new output value is yielded. This process can be schematized as follows: 

\begin{equation}\label{eq:reduce}
Reduce(<key2, list(val2)>) \rightarrow <key3, val3>
\end{equation} 



Apache Hadoop~\cite{whi12,had15} is an open-source implementation of MapReduce for reliable, scalable, distributed computing. Despite being the most popular open-source implementation of MapReduce, Hadoop is not suitable in many cases, such as online and/or iterative computing, high inter-process communication paradigms or in-memory computing, among others~\cite{lin12}. 

In recent years, Apache Spark has been introduced in the Hadoop Ecosystem~\cite{hamstra15,spa15}. This powerful framework is aimed at performing faster distributed computing on Big Data by using in-memory primitives that allows it to perform 100 times faster than Hadoop for certain applications. This platform allows user programs to load data into memory and query it repeatedly, making it a well suited tool for online and iterative processing (especially for machine learning algorithms). Additionally, it provides a wider range of primitives that ease the programming task. 

Spark is based on a distributed data structures called Resilient Distributed Datasets (RDDs). By using RDDs, we can implement several distributed programming models like Pregel or MapReduce, thanks to their generality capability. These parallel data structures also let programmers persist intermediate results in memory and manage the partitioning to optimize data placement.

As a subproject of Spark, a scalable machine learning library (MLlib)~\cite{mll15} was created. MLlib is formed by common learning algorithms and statistic utilities. Among its main functionalities includes: classification, regression, clustering, collaborative filtering, optimization, and dimensionality reduction (mostly feature extraction).

\section{Filtering Feature Selection for Big Data}
\label{sec:itfsbig}

In~\cite{brown12}, an Information Theoretic framework that includes many common FS filter algorithms was proposed. In their work the authors prove that algorithms like minimum Redundancy Maximum Relevance (mRMR) and others are special cases of Conditional Mutual Information when some specific independence assumptions are made about both the class and the features (see details below). Here, we demonstrate that these criteria are not only a sound theoretical formulation, but it also fits well in modern Big Data platforms and allows us to distribute several FS methods and their complexity across a cluster of machines. 

In this work, we describe how we have redesigned this framework for a distributed paradigm. This version contains a generic implementation of several Information Theoretic FS methods such as: mRMR, Conditional Mutual Information Maximization (CMIM), or Joint Mutual Information (JMI), among others; that furthermore have been designed to be integrated in the MLlib Spark library. Additionally, the framework can be extended with other criteria provided by the user as long as they comply with the guidelines proposed by Brown \textit{et al.}


In Section~\ref{subsec:itheory}, we present this framework and explain the process to adapt it to the Big Data environment. Section~\ref{subsec:framework} describes how the selection process and the underlying information theory operations have been implemented in a distributed manner by using Spark primitives.
 
\subsection{Filter methods based on Information Theory}
\label{subsec:itheory}


Information measures tell us how much information has been acquired by the receiver when he/she gets a message~\cite{mark73}. In predictive learning, we associate the message with the output feature in classification. 

A commonly used uncertainty function is Mutual Information (MI)~\cite{cover91}, which measures the amount of information one random variable contains about another. This is, the reduction in the uncertainty of one random variable due to the knowledge of the other:

\begin{equation}
\label{eq:mi}
\begin{aligned}
I(A;B) &= H(A) - H(A|B) \\
       &= \sum_{a\in A} \sum_{b\in B} p(a,b) \log \frac{p(a,b)}{p(a) p(b)}.
\end{aligned}
\end{equation}

\noindent where $A$ and $B$ are two random variables with marginal probability mass functions $p(a)$ and $p(b)$, respectively; $p(a,b)$ the joint mass function and $H$ the entropy.


In the same way, MI can be conditioned to a third random variable. Thus, Conditional Mutual Information (CMI) is denoted as:

\begin{equation}
\label{eq:cmi}
\begin{aligned}
I(A;B|C) &= H(A|C) - H(A|B,C)\\
         &= \sum_{c\in C} p(c) \sum_{a\in A} \sum_{b\in B} p(a,b,c) \log \frac{p(a,b,c)}{p(a,c) p(b,c)}.
\end{aligned}
\end{equation}
\noindent where $C$ is a third random variable with marginal probability mass function $p(c)$; and $p(a,c)$, $p(b,c)$ and $p(a,b,c)$ the joint mass functions.

%

%
%

Filtering methods are based on a quantitative criterion or index, also known as relevance index or scoring. This index is aimed at measuring the usefulness of each feature for a specific classification problem. Through the~\textbf{relevance} (self-interaction) of a feature with the class, we can rank the features and select the most relevant ones.
However, the features can also be ranked using a more complex criterion such as if a feature is more~\textbf{redundant} than another (multi-interaction). For instance, redundant features can be discarded (those variables that carry similar information) using the Mutual Information criterion~\cite{battiti1994using}:

$$J_{mifs}(X_i) = I(X_i; Y) - \beta \sum_{X_j \in S} I(X_i; X_j),$$

\noindent where $S \subseteq S_\theta$ is the current set of selected features and $\beta$ is a weight factor. It considers the MI between each candidate $X_i \not\in S$ and the class, but also introduces a penalty proportional to its redundancy, calculated as the MI between the current set of selected features and each candidate. 

There are a wide range of methods in the literature built on these information theoretic measures. To homogenize the use of all these criteria, Brown \textit{et al.}~\cite{brown12} proposed a generic expression that allows to ensemble multiple information theoretic criteria into a unique FS framework. This framework is based on a greedy optimization process which assesses features based on a simple scoring criterion. Through some independence assumptions, it allows to transform many criteria as linear combinations of Shannon entropy terms: MI and CMI~\cite{cover91}. In some cases, it expresses more complex criteria as non-linear combinations of these terms (e.g. max or min). For a detailed description of the transformation processes, please see~\cite{brown12}. The generic formula proposed by Brown~\textit{et al.}~\cite{brown12} is:

\begin{equation}
\label{eq:gen_form}
J = I(X_i;Y) - \beta \sum_{X_j \in S} I(X_j;X_i) + 
\gamma \sum_{X_j \in S} I(X_j;X_i|Y),
\end{equation}

\noindent where $\gamma$ represents a weight factor for the conditional redundancy part.

The formula can be divided into three parts: the first one represents the relevance of a feature $X_i$, the second one the redundancy between two features $X_i$ and $X_j$, and the last one the conditional redundancy between two features $X_i, X_j$ and the class $Y$. Through the aforementioned assumptions, many criteria were re-written by the authors to fit the generic formulation so that all these methods could be implemented with a slight variation in this formula. In Table~\ref{tab:criterios}, we show a comprehensive list of methods implemented in our proposal according to the adaptation proposed by Brown \textit{et al.}

\begin{table*}[!htbp]\renewcommand{\arraystretch}{1.3}
\small
\centering
\resizebox{\textwidth}{!}{
\begin{tabular}{|l|l|}
\hline
\multicolumn{2}{|l|}{\textbf{Criterion name}} \\
\hline
\textbf{Original proposal} & \textbf{Brown's reformulation} \\
\hline \hline
\multicolumn{2}{|l|}{\textit{Mutual Information Maximisation} (MIM)~\cite{lewis92}} \\ \hline
$J_{mim}(X_i) = I(X_i; Y)$ &
$J_{mim} = I(X_i;Y) - 0 \sum_{X_j \in S} I(X_j;X_i) + 0 \sum_{X_j \in S} I(X_j;X_i|Y)$ \\
\hline\hline
\multicolumn{2}{|l|}{\emph{Mutual Information FS} (MIFS)~\cite{battiti1994using}} \\ \hline
$J_{mifs}(X_i) = I(X_i;Y) - \beta \sum_{X_j\in S} I(X_i;X_j)$ &
$J_{mifs} = I(X_i;Y) - \beta \sum_{X_j \in S} I(X_j;X_i) + 0 \sum_{X_j \in S} I(X_j;X_i|Y)$ \\
\hline\hline
\multicolumn{2}{|l|}{\textit{Joint Mutual Information} (JMI)~\cite{yang1999}} \\ \hline
$J_{jmi}(X_i) = \sum_{X_j \in S} I(X_iX_j;Y)$ &
$J_{jmi} = I(X_i;Y) - \frac{1}{|S|} \sum_{X_j \in S} I(X_j;X_i) + \frac{1}{|S|} \sum_{X_j \in S} I(X_j;X_i|Y)$ \\
\hline\hline
\multicolumn{2}{|l|}{\textit{Conditional Mutual Information} (CMI)} \\ \hline
$J_{cmi} = I(X_i;Y|S)$ &
$J_{cmi} = I(X_i;Y) - \sum_{X_j \in S} I(X_j;X_i) + \sum_{X_j \in S} I(X_j;X_i|Y)$ \\
\hline\hline
\multicolumn{2}{|l|}{\textit{Minimum-Redundancy Maximum-Relevance} (mRMR)~\cite{peng2005}} \\ \hline
$J_{mrmr} = I(X_i;Y) - \frac{1}{|S|} \sum_{X_j \in S} I(X_j;X_i)$ &
$J_{mrmr} = I(X_i;Y) - \frac{1}{|S|} \sum_{X_j \in S} I(X_j;X_i) + 0 \sum_{X_j \in S} I(X_j;X_i|Y)$ \\
\hline\hline
\multicolumn{2}{|l|}{\textit{Conditional Mutual Information Maximization} (CMIM)~\cite{fleuret2004}} \\ \hline
$J_{cmim} = \min_{X_j \in S} [I(X_i;Y|X_j)]$ &
$J_{cmim} = I(X_i;Y) - \max_{X_j \in S}[I(X_j;X_i) - I(X_j;X_i|Y)]$ \\
\hline\hline
\multicolumn{2}{|l|}{\textit{Informative Fragments} (IF)~\cite{vidal2003} (equivalent to CMIM)} \\ \hline
$J_{if} = \min_{X_j \in S} [I(X_iX_j;Y) - I(X_j;Y)]$ &
$J_{if} = J_{cmim} = I(X_i;Y) - \max_{X_j \in S}[I(X_j;X_i) - I(X_j;X_i|Y)]$ \\
\hline\hline
\multicolumn{2}{|l|}{\textit{Interaction Capping} (ICAP)~\cite{tesisJakulin}}\\ \hline
$J_{icap} = I(X_i;Y) - \sum_{X_j \in S} \max [0, I(X_i;X_j) - I(X_i;X_j|Y)]$ &
$J_{icap} = I(X_i;Y) - \sum_{X_j \in S} \max [0, I(X_i;X_j) - I(X_i;X_j|Y)]$ \\
\hline
\end{tabular}
}
\caption{Implemented Information Theoretic criteria: originals and adaptations.}
\label{tab:criterios}
\end{table*}
 

\subsection{Filter FS Framework for Big Data}
\label{subsec:framework}

Here, we present the proposed FS framework for Big Data using distributed operations. We outline the most important improvements carried out to adapt the classical approach to this new Big Data environment. Similarly, we analyze the implications derived from the distributed implementation of Equation~\ref{eq:gen_form}, as well as the complexity derived from the parallelization of the core operations of this expression: MI and CMI. 

Beyond the implementation on Spark, we have re-designed Brown's framework by adding some new important improvements to the performance of the classical approach, but also maintaining some features of this one: 

\begin{itemize}

\item \textbf{Columnar transformation:} The access pattern presented by most FS methods is thought to be feature-wise; in contrast to many other ML algorithms, which are used to work with rows (instance-wise). Despite being a simple detail, this can significantly degrade the performance since the natural way of computing relevance and redundancy in FS methods is normally thought to be performed by columns. This is specially important for distributed frameworks like Spark, where the partitioning scheme of data is quite influential in the performance.

\item \textbf{Use of broadcasting}: Once all features values are grouped and partitioned into different partitions, minimum data shift should occurred in order to avoid superfluous network and CPU usage. So if the MI process is performed locally in each partition, the overall algorithm will run efficiently (almost linearly) . We propose to minimize the data movement by replicating the output feature and the last selected feature in each iteration.

\item \textbf{Caching pre-computed data:} The first term that appears in the generic criterion of Equation~\ref{eq:gen_form} is relevance, which basically implies to calculate MI between all input features and the output (relevance). This operation is performed once at the start of our algorithm, then cached to be re-used in the next evaluations of Equation~\ref{eq:gen_form}. Likewise, the subsequent marginal and joint proportions derived from these operations are also kept to omit some computations. This will also help to isolate the computation of redundancy per feature by replicating this permanent information in all nodes. 

\item \textbf{Greedy approach:} Brown \textit{et al.} proposed a greedy search process so that only one feature is selected in each iteration. This fact transform the quadratic complexity of typical FS algorithms into a more manageable complexity determined by the number of features to select.
\end{itemize}

We have also employed some complex operations from Spark API, which we present below. Spark primitives extend the idea of MapReduce to offer much more complex operations that ease code parallelization. Here, we outline those more relevant for our method\footnote{For a complete description of Spark's operations, please refer to Spark's API: \url{https://spark.apache.org/docs/latest/api/scala/index.html}}:

\begin{itemize}
\item $mapPartitions$: Similar to Map, this runs a function independently on each partition. For each partition, an iterator of tuples is fetched and another of the same type is generated.
\item $groupByKey$: This operation groups those tuples with the same key in a single vector of values (using a shuffle operation).
\item $sortByKey$: A distributed version of merge sort.
\item $broadcast$: This operation allows to keep a read-only copy of a given variable on each node rather than shipping a copy to each task. This is normally used for large permanent variables (such as big hash tables).
\end{itemize}


\subsubsection{Main FS Algorithm}
\label{subsubsec:main}

In Algorithm~\ref{alg:main}, the main algorithm for selecting features is presented. This procedure is in charge of deciding which feature to select in a sequential manner. Roughly, it calculates the initial relevances for all the features, and iterates over it selecting the best features according to Equation~\ref{eq:gen_form} and the underlying MI and CMI values.

\begin{algorithm}[!htb]
\small
\caption{Main FS Algorithm}
\label{alg:main}
\begin{algorithmic}
\INPUT $D$ Dataset, an RDD of samples.
\INPUT $ns$ Number of features to select.
\INPUT $npart$ Number of partitions to set.
\INPUT $cindex$ Index of the output feature.
\OUTPUT $S_\theta$ Index list of selected features
\STATE $D_c \leftarrow columnarTransformation(D, ns, npart)$
\STATE $ni \leftarrow D.nrows; nf \leftarrow D.ncols$
\STATE $REL \leftarrow computeRelevances(D_c, cindex, ni)$
\STATE $CRIT \leftarrow initCriteria(REL)$
\STATE $p_{best} \leftarrow CRIT.max$
\STATE $sfeat \leftarrow Set(p_{best})$
\WHILE{$|S| < |S_\theta|$}
     \STATE $RED \leftarrow computeRedundancies(D_c, p_{best}.index)$
     \STATE $CRIT \leftarrow updateCriteria(CRIT, RED)$
     \STATE $p_{best} \leftarrow CRIT.max$
     \STATE $sfeat \leftarrow addTo(p_{best}, sfeat)$
\ENDWHILE
\STATE $return(sfeat)$
\end{algorithmic}
\end{algorithm}

The first step consists of transforming data into a columnar format as proposed in the list of improvements. Once data matrix is transformed, the algorithm obtains the relevance for each feature in $X$, initializing the criterion value (partial result according to Equation~\ref{eq:gen_form}), and creates an initial ranking of the features. Relevance values are saved as part of the previous expression and re-used in next steps to update the criteria. Afterwards, the most relevant feature, $p_{best}$, is selected and added to the set $sfeat$, which is empty at first. The iterative phase begins by calculating MI and CMI between $p_{best}$, each candidate $X_i$, and $Y$. The subsequent values will serve to update the accumulated redundancies (simple and conditional) of the criteria. At each iteration, the most relevant candidate feature will be selected as the new $p_{best}$ and added to $sfeat$. The loop ends when $ns$ features (where $ns = |S_\theta|$) have been selected, or there are no more features to select.



\subsubsection{Distributed Operations: Columnar Transformation and MI Computations}
\label{subsubsec:distributed}

The estimation of MI and CMI are undoubtedly the costliest operations in Information Theoretic FS. When we face huge datasets, these operations are sequentially unfeasible to calculate as the number of combinations grow. This section describes how these calculations have been parallelized towards a set of distributed operations (explained in Section~\ref{subsec:mapreduce}). For all the algorithms described below, RDD variables have been highlighted in uppercase in order to differentiate them from the ordinary variables. 

\textbf{Columnar Transformation}

Columnar format is clearly much more manageable for filter FS methods than row-wise format, as mentioned before. 
Algorithm~\ref{alg:columnar} explains this transformation, carried out in our algorithm as the the first step. The idea behind this transformation is to transpose the local data matrix provided by each partition. This partition operation will maintain the partitioning scheme without incurring in a high shuffling overhead. Additionally, once data are transformed, they can be cached and re-use in the subsequent loop. The result of this operation is a new matrix with one row per feature. It generates a tuple, where $k$ represents the feature index, $part.index$ the index of the partition (henceforth block index) and $matrix(k)$ the local matrix for this feature-block.

\begin{algorithm}[!htb]
\small
\caption{Function that transform row-wise data into a columnar format~\textit{(columnarTransformation)}}
\label{alg:columnar}
\begin{algorithmic}[1]
\INPUT $D$ Dataset, an RDD of samples.
\INPUT $nf$ Number of features.
\INPUT $npart$ Number of partitions to set.
\OUTPUT Column-wise data (RDD of feature vectors).
\STATE $D_c \leftarrow $ \MAPP{$part \in D$}
	\STATE $matrix \leftarrow new~Matrix(nf)(part.length)$
	\FOR{$j=0~until~part.length$}
		\FOR{$i=0~until~nf$}
		\STATE $matrix(i)(j) \leftarrow part(j)(i)$
		\ENDFOR
	\ENDFOR
	\FOR{$k=0~until~nf$}
		\STATE $EMIT <k, (part.index, matrix(k))>$
	\ENDFOR
\ENDMAPP
\STATE $return(D_c.sortByKey(npart))$
\end{algorithmic}
\end{algorithm}

In order to benefit from data locality, the algorithm allocates all instances of the same feature in a determined set of partitions (if possible, only in one). To do that, this sorts the new instances by key, limiting the number of partitions to $npart$. In next phases, the partitions will be mapped with the aim of generating a number of histograms per feature, which count the number of occurrences by combination. 

Choosing a proper number of partitions is important for the next steps. If $npart$ is equal or less than the number of features, the number of total histograms per feature will be two at most. On the contrary, if this number is greater than the number of features, the total number of histograms generated per feature can be high since the same feature can be distributed across many partitions (more than two). We thus recommend setting this parameter to $2\times$ the number of features at most.

\begin{figure*}[!htp]
\centering
\includegraphics[width=0.75\textwidth]{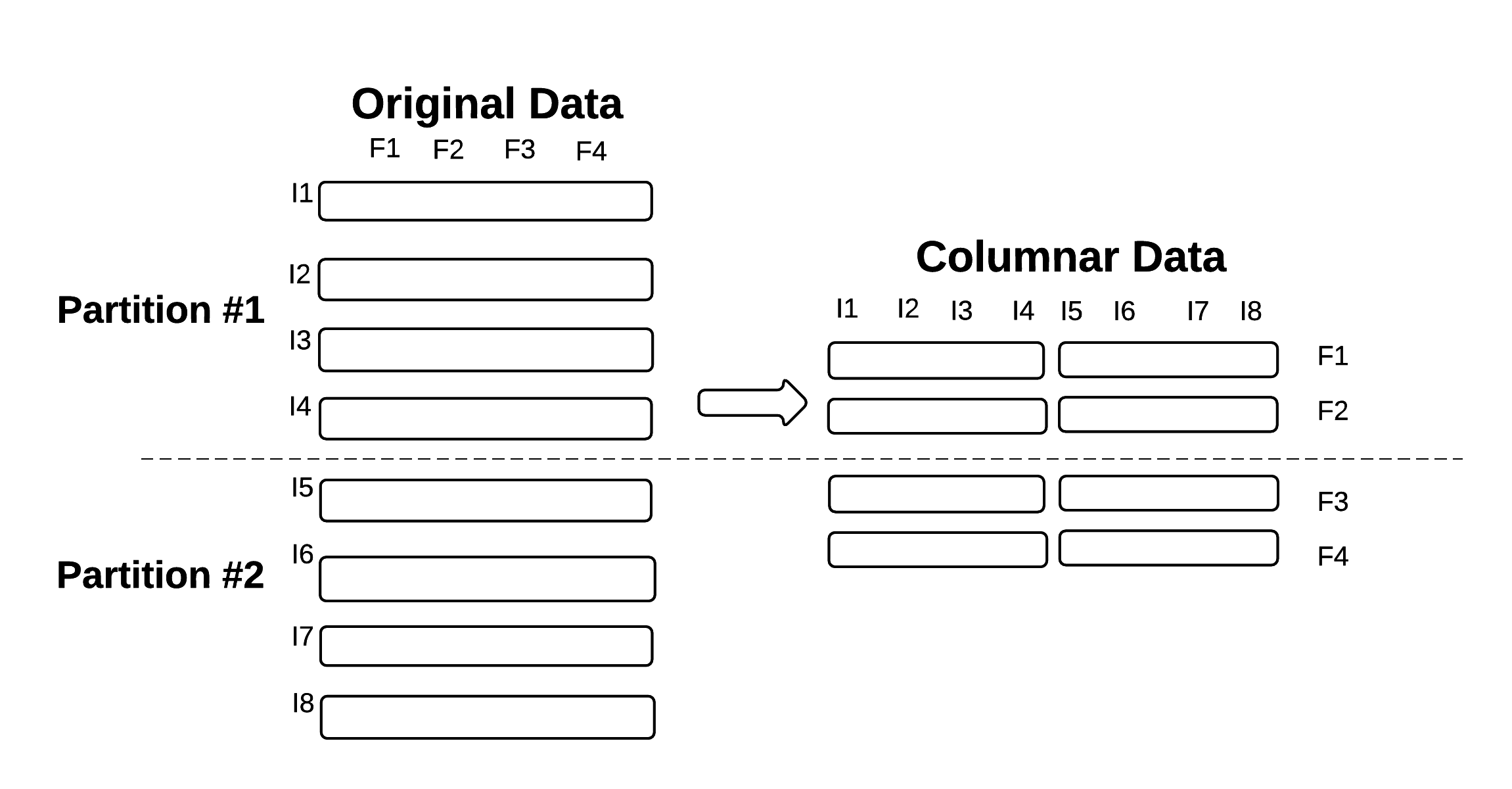}
\caption{Columnar transformation scheme. \textit{F} indicates features and \textit{I} instances. Each rectangle on the left represents a single register in the original dataset. Each rectangle on the right represents a transposed feature block in the new columnar format.}
\label{fig:columnar}
\end{figure*}

Figure~\ref{fig:columnar} details this process using a small example with eight instances and four features. In this figure, we can see how the algorithm generates a block for each feature in each partition. Then, all blocks are sorted by feature in order to gather them in the same partitions.

\textbf{Computing Relevance}

After transforming data, Algorithm~\ref{alg:relevances} describes how to compute relevance (MI) between all the input features and $Y$ (as expressed in Equation~\ref{eq:mi}). This has been designed as an initialization method, so that all variables that appear in this function can be used in the subsequent algorithms. For example, the number of distinct values for each feature is first computed and saved as $counter$ (to limit the size of histograms). 

\begin{algorithm}[!htb]
\small
\caption{Compute mutual information between the set of features $X$ and $Y$.~\textit{(computeRelevances)}}
\label{alg:relevances}
\begin{algorithmic}[1]
\INPUT $D_c$ RDD of tuples (index, (block, vector)).
\INPUT $yind$ Index of $Y$.
\INPUT $ni$ Number of instances.
\OUTPUT MI values for all input features.
\STATE $ycol \leftarrow D_c.lookup(yind)$
\STATE $bycol \leftarrow broadcast(ycol)$
\STATE $counter \leftarrow broadcast(getMaxByFeature(D_c))$
\STATE $H \leftarrow getHistograms(D_c, yind, bycol, null, null)$
\STATE $joint \leftarrow getProportions(H, ni)$
\STATE $marginal \leftarrow getProportions(aggregateByRow(joint), ni)$
\STATE $return(computeMutualInfo(H, yind, null))$
\end{algorithmic}
\end{algorithm}

The main idea behind relevance and redundancy functions is to perform the calculations for each feature independently. This is done by distributing only the single variables ($p_{best}$ and $Y$) across the cluster, and leverage for the aforementioned data locality property. In this case, the first step consists of collecting all blocks of $Y$ from the data, and putting all of them in a single vector to be broadcasted ($bycol$). Histograms for all the candidate features with respect to $Y$ are then calculated in $getHistograms$ (explained below). This function is common to the relevance and redundancy phases. This computes 3-dimensional histograms between all non-selected features and two secondary variables (for redundancy) and between all non-selected features and one variable (for relevance)\footnote{For relevance, the~\textit{null} value is used to represent the lack of the second variable}. Joint and marginal proportions are generated from the resulting histograms using matrix operations: aggregating proportions by row ($marginal$), and computing proportions for joint ($joint$). Finally, using this information we can now obtain the MI value for each candidate feature.

\textbf{Computing Redundancy}

In this case, the computation of the simple and conditional redundancy is performed between $p_{best}$, each candidate feature $X_i$ and $Y$. The conditional redundancy introduces a third conditional variable ($Y$), following the formula: $I(X_j; X_i | Y)$ (introduced in Equation~\ref{eq:cmi}).

This operation is repeated until we reach the number of selected features that is specified as a parameter. Algorithm~\ref{alg:redundancies} details this process, which is an extension of relevance computation (Algorithm~\ref{alg:relevances}). This obtains the blocks for $p_{best}$ from the RDD, and broadcasts them to all the nodes. Then, the function $getHistograms$ is called with two variables in order to obtain the histograms for all the candidate features with respect to $p_{best}$, and $Y$. Note that the vector for $Y$ is already available from the redundancy phase. Finally, both types of redundancy are computed using the function that computes MI and CMI ($computeMutualInfo$).

\begin{algorithm}[!htb]
\small
\caption{Compute CMI and MI between $p_{best}$, the set of candidate features, and $Y$.~\textit{(computeRedundancies)}}
\label{alg:redundancies}
\begin{algorithmic}[1]
\INPUT $D_c$ RDD of tuples (index, (block, vector)).
\INPUT $jind$ Index of $p_{best}$.
\OUTPUT CMI values for all input features.
\STATE $jcol \leftarrow D_c.lookup(jind)$
\STATE $bjcol \leftarrow broadcast(jcol)$
\STATE $H \leftarrow getHistograms(D_c, jind, bjcol, yind, bycol)$
\STATE $return(computeMutualInfo(H, jind, yind))$
\end{algorithmic}
\end{algorithm}

\textbf{Histograms creation}

Algorithm~\ref{alg:hist} computes 3-dimensional histograms for the set of candidate features with respect to $p_{best}$ and $Y$, which later will be used to compute MI and CMI. In case of not providing the conditional variable, this yields histograms whose third dimension is equal to one. 

\begin{algorithm}[!htb]
\small
\caption{Function that computes 3-dimensional histograms between $p_{best}$, the set of candidate features, and $Y$ for CMI; or between the set of candidate features, and $Y$ for MI.~\textit{(getHistograms)}}
\label{alg:hist}
\begin{algorithmic}[1]
\INPUT $D_c$ RDD of tuples (index, (block, vector)).
\INPUT $jind$ Index of $Y$ or $p_{best}$.
\INPUT $yind$ Index of feature $Y$ (can be empty).
\INPUT $jcol$ Values for $Y$ or $p_{best}$, a broadcasted matrix.
\INPUT $ycol$ Values for $Y$, a broadcasted matrix (can be empty).
\OUTPUT Columnar-wise dataset (RDD of feature vectors).
\STATE $jsize \leftarrow counter(jind)$
\STATE $ysize \leftarrow counter(yind)$
\STATE $H \leftarrow $\MAPP{$part \in partitions$}
	\FOR{$(k, (block, v)) \leftarrow part$}
		\STATE $isize \leftarrow counter(k)$
		\STATE $m \leftarrow new Matrix(ysize)(isize)(jsize)$
		\FOR{$e=0~until~v.size$}
			\STATE $j \leftarrow jcol(block)(e);~y \leftarrow ycol(block)(e);~i \leftarrow v(e)$
			\STATE $m(y)(i)(j) += 1$
		\ENDFOR
		\STATE $EMIT <k, m>$
	\ENDFOR
\ENDMAPP
\STATE $return(H.reduceByKey(sum))$
\end{algorithmic}
\end{algorithm}

The first two lines return the dimensions for $p_{best}$ and $Y$ variables using $counter$ (Algorithm~\ref{alg:main}). Then, a map operation on each partition is started on the dataset. This operation iterates over the blocks derived from the columnar transformation. Each tuple is formed by a key (the index of a candidate feature), and a value with an index block and the corresponding feature array ($(k, (block, v))$). For each instance, a matrix is initialized to zero and then incremented by one depending on the value in each combination $p_{best}$, $X_i$, and $Y$. Single features ($p_{best}$ and $Y$) are broadcasted in form of matrices, whose first axis indicates the block index, and the second one the partial index of the value in this block. This updating operation is repeated until the end of the feature vector. After that, a new tuple is emitted with the feature index as key and the resulting matrix as value ($<k, m>$). The map operation continues with the next block until finishing the partition. Finally, final histograms are aggregated by summing them up. This aggregation process will remain simple as long as the number of histograms per partition is small. This is normally true as a single partition usually contains all the blocks for the same feature. Therefore, it is important to reduce the number of histograms by adjusting the number of partitions, as explained above.

\begin{figure*}[!htp]
\centering
\includegraphics[width=0.75\textwidth]{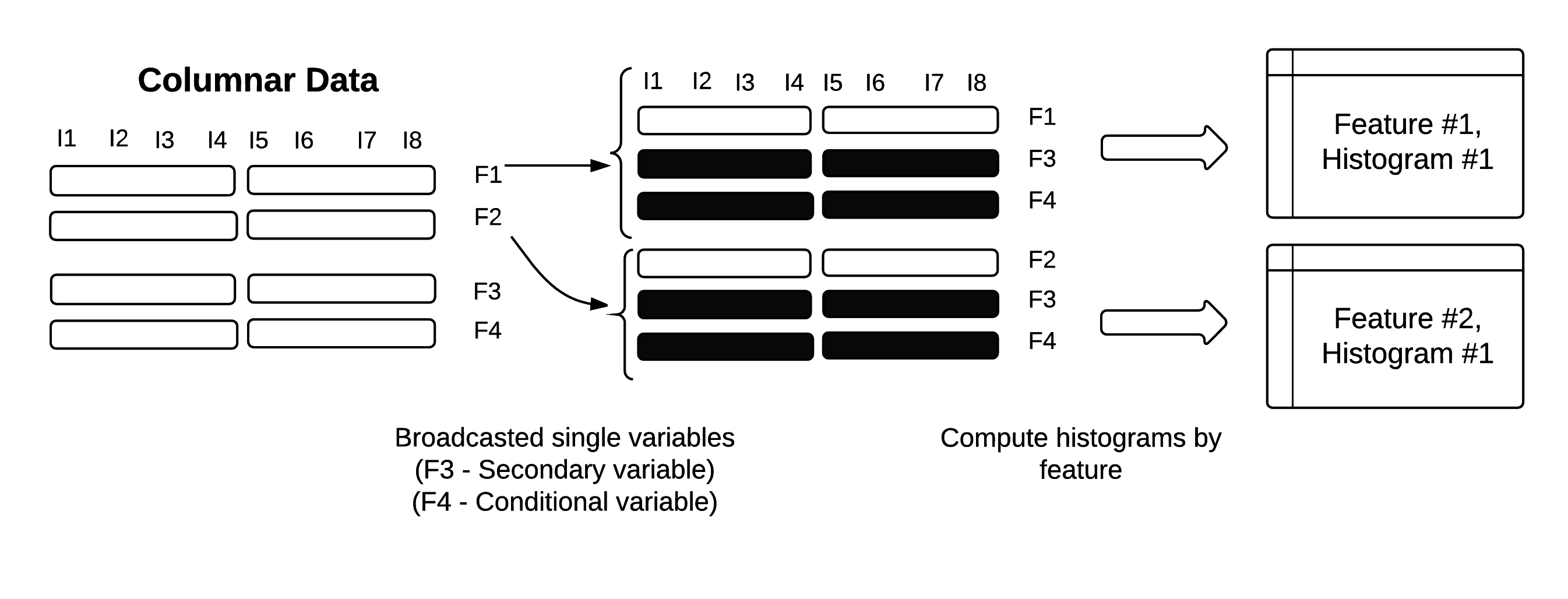}
\caption{Histograms creation scheme. \textit{F} indicates features and \textit{I} instances. Each white rectangle represents a single feature block in the columnar format. Black rectangles represent marginal and joint proportions for single variables, which are broadcasted across the cluster.}
\label{fig:histograms}
\end{figure*}

Figure~\ref{fig:histograms} details the histograms creation process using a simple example. In this figure, we can observe how the algorithm generates one histogram for each partition and feature. In this case, the number of partitions corresponds with the number of features, so only one histogram per feature is generated. 

\textbf{MI and CMI computations}

Algorithm~\ref{alg:minfo} details the process that unifies the computation of MI and CMI. The algorithm takes as input the indices of single variables ($Y$ or $p_{best}$ for MI, and both variables for CMI), and all previously computed histograms. Before starting, the algorithm broadcasts the marginal and joint matrices that correspond to these variables. All this information is sent to the nodes because this cannot be computed from the previous histograms independently, and is already computed.

\begin{algorithm}[!htb]
\small
\caption{Calculate MI and CMI for the set of histograms with respect to $Y$ or $p_{best}$ for MI, and both variables for CMI.~\textit{(computeMutualInfo)}}
\label{alg:minfo}
\begin{algorithmic}[1]
\INPUT $H$ Histograms, an RDD of tuples (index, matrix).
\INPUT $bind$ Index of feature $p_{best}$.
\INPUT $cind$ Index of feature $Y$ (can be empty).
\OUTPUT MI and CMI values.
\STATE $bprob \leftarrow broadcast(marginal(bind))$
\STATE $cprob \leftarrow broadcast(marginal(cind))$
\STATE $bcprob \leftarrow broadcast(joint(bind))$
\STATE $MINFO \leftarrow $\MAP{$(k, m) \in H$}
	\STATE $aprob \leftarrow computeMarginal(m)$
	\STATE $abprob \leftarrow computeJoint(m, bind);~acprob \leftarrow computeJoint(m, cind)$
	\FOR{$c=0~until~getSize(m)$}
		\FOR{$b=0~until~getnRows(m(c))$}
			\FOR{$a=0~until~getnCols(m(c))$}
				\STATE $pc \leftarrow cprob(c);~pabc \leftarrow (m(c)(a)(b) / ninstances) / pc$
				\STATE $pac \leftarrow acprob(c)(a);~pbc \leftarrow bcprob(c)(b)$
				\STATE $cmi~+= conditionalMutualInfo(pabc, pac, pbc, pc)$
				\IF{$c == 0$}
					\STATE $pa \leftarrow xprob(a);~pab \leftarrow abprob(a)(b);~pb \leftarrow yprob(b)$
					\STATE $mi~+= mutualInfo(pa, pab, pb)$
				\ENDIF
			\ENDFOR
		\ENDFOR
	\ENDFOR
	\STATE $EMIT <k, (mi, cmi)>$
\ENDMAP
\STATE $return(MINFO)$
\end{algorithmic}
\end{algorithm}

A map phase is then started on each histogram tuple, which consists of a given feature index as key and a 3-dimensional matrix as value. In this phase, the algorithm generates the MI and CMI values for all the combinations between the histograms and the single variables (see Equations~\ref{eq:mi} and~\ref{eq:cmi}). Before that, it is needed to compute the marginal proportions for the set of candidate features, and the joint proportions between each $X_i$ and $Y$; and, each $X_i$ and $p_{best}$ (using matrix operations as described in~\ref{alg:relevances}). Once all joint and marginal proportions are calculated, a loop starts over all combinations to compute the proportion $pabc$ (which comes directly from the histogram), and finally, the final result by combination. All these results are then aggregated to get the overall MI and CMI values for each feature.

\begin{figure*}[!htp]
\centering
\includegraphics[width=0.75\textwidth]{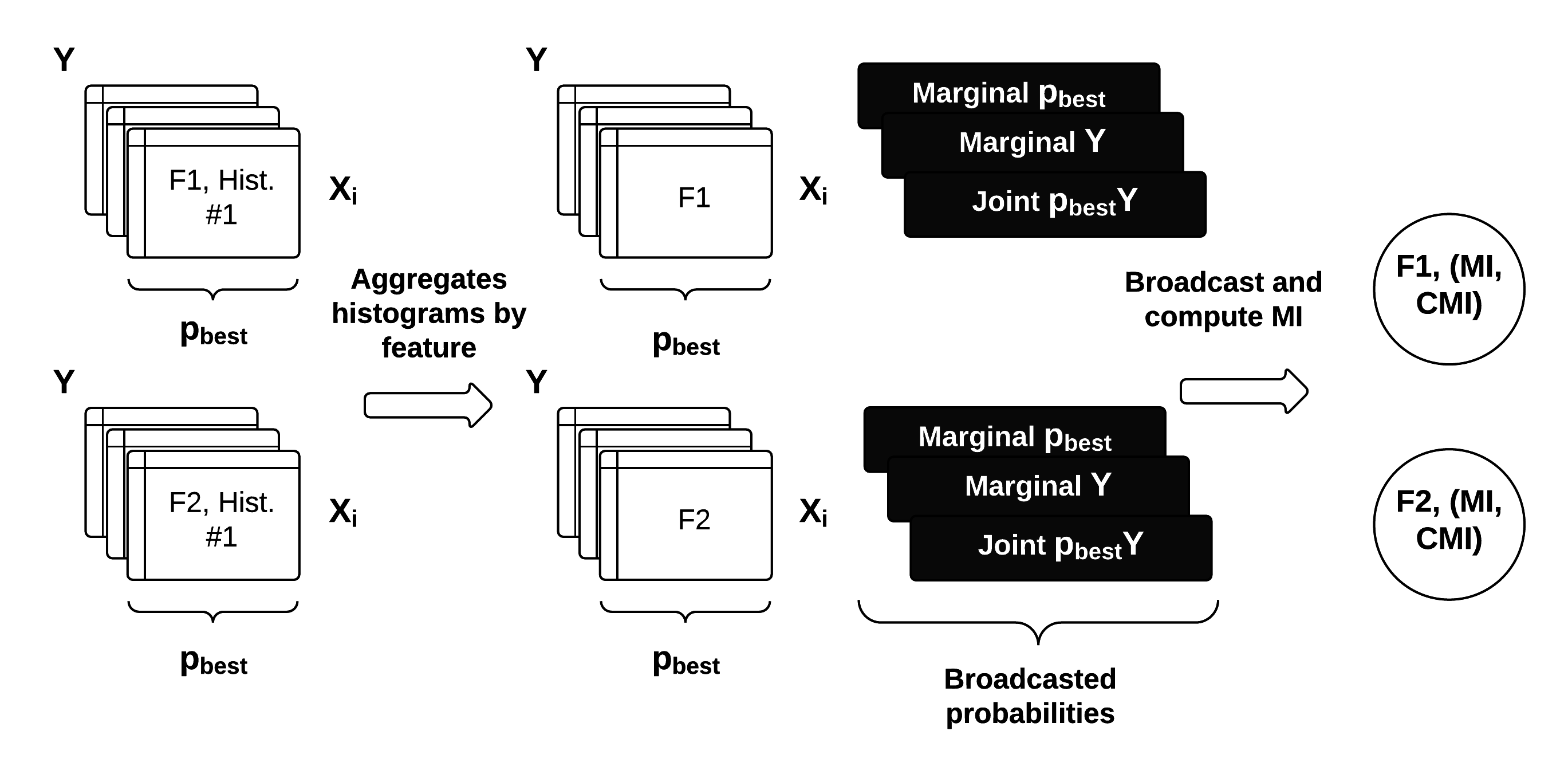}
\caption{MI computations. \textit{F} indicates features and the matrices indicates the histograms for each feature. Black rectangles represents broadcasted joint and marginal probabilities.}
\label{fig:mutualinfo}
\end{figure*}

Figure~\ref{fig:mutualinfo} details the MI process started once all histograms have been computed. Each partition generates a histogram for each feature contained in it. Histograms for the same feature are aggregated to obtain a single final histogram. Only those marginal and joint proportions that can not be computed independently from each histograms are broadcasted. By using the histograms and these broadcasted variables, MI and CMI are computed per feature in an independent way. 

\subsubsection{High-dimensional and sparse version}
\label{subsubsec:sparse}

Previous version works well with ``tall and skinny" data, formed by a small number of features and a large number of instances. However, when we face datasets with millions of features and with a high sparsity, this implies a new problem where the complexity grows in the opposite size (the horizontal one). It is therefore mandatory to re-adapt the previous algorithms in order to deal with such amount of characteristics, thus preventing an undermined performance. In this case, the algorithms highly affected by this change are detailed: the columnar transformation (Algorithm~\ref{alg:columnar}) and the histogram creation process (Algorithm~\ref{alg:hist}). The rest of code remains unchanged except the structure of data, which is reduced to a single vector (in form of $(index, vector)$). The block index is therefore removed from this structure as only one histogram is generated per each feature.

As explained before, high-dimensional sparse data is usually characterized by a large number of features and a variable number of instances with an undefined number of non-zero indexed elements. 
Algorithm~\ref{alg:colsparse} presents a new model of processing that transposes data directly, generating single vectors for each feature. Sparsity is maintained on sparse vectors but employing the index of the original instance as key. The algorithm generates a tuple for each value so that the key is formed by the feature index, and the value is formed by the instance index and the value itself. All tuples are grouped by key to create a single sparse vector (formed by sorted key-value tuples).

\begin{algorithm}[!htb]
\small
\caption{Function that transforms row-wise data into columnar data (sparse version)~\textit{(sparseColumnar)}}
\label{alg:colsparse}
\begin{algorithmic}[1]
\INPUT $D$ Dataset, an RDD of sparse samples.
\INPUT $npart$ Number of partitions to set.
\OUTPUT Column-wise dataset (feature vectors).
\STATE $D_c \leftarrow $ \MAP{$reg \in D$}
	\STATE $ireg \leftarrow reg.index$
	\FOR{$i=0~until~reg.length$}
		\STATE $EMIT <reg(i).key, (ireg, reg(i).value)>$
	\ENDFOR
\ENDMAP
\STATE $D_c \leftarrow D_c.groupByKey(npart).mapValues(vectorize)$
\end{algorithmic}
\end{algorithm}

Regarding the histogram creation, in this new version the map partition operation has been replaced by a map operation, which is applied to each feature vector previously generated. Algorithm~\ref{alg:histsparse} details this process. This is quite similar to Algorithm~\ref{alg:hist} with the caveat that only one histogram is yielded for each feature and, therefore, no reduce operation is needed. 

\begin{algorithm}[!htb]
\small
\caption{Function that computes 3-dimensional histograms for the set of features $X_i$ with respect to features $p_{best}$ and $Y$ (sparse version).~\textit{(sparseHistograms)}}
\label{alg:histsparse}
\begin{algorithmic}[1]
\INPUT $D_c$ Dataset, an RDD of tuples (Int, (Int, Vector)).
\INPUT $jind$ Index of $Y$ or $p_{best}$.
\INPUT $yind$ Index of feature $Y$ (can be empty).
\INPUT $jcol$ Values for $Y$ or $p_{best}$, a broadcasted matrix.
\INPUT $ycol$ Values for $Y$, a broadcasted Matrix (can be empty).
\OUTPUT Columnar-wise dataset (RDD of feature vectors).
\STATE $jsize \leftarrow counter(jind);~ysize \leftarrow counter(yind)$
\STATE $jyhist \leftarrow frequencyMap(jind, yind)$
\STATE $zhist \leftarrow frequencyMap(yind)$
\STATE $H \leftarrow $\MAP{$(k, v) \in D_c$}
		\STATE $isize \leftarrow counter(k)$
		\STATE $m \leftarrow new Matrix(ysize)(isize)(jsize)$
		\FOR{$e=0~until~v.size$}
			\STATE $j \leftarrow jcol(e);~y \leftarrow ycol(e)$
			\STATE $i \leftarrow v(e)$
			\IF{$j <> 0$}
				\STATE $jyhist(j)(y) = jyhist(j)(y) - 1$
			\ENDIF
			\STATE $m(y)(i)(j) += 1$
		\ENDFOR
		\FOR{$((j, y), q) \leftarrow jyhist$}
			\STATE $m(y)(0)(j) += q$
		\ENDFOR
		\FOR{$(y, q) \leftarrow yhist$}
			\STATE $m(y)(0)(0) += yhist(y) - sum(mat(y))$
		\ENDFOR		
		\STATE $EMIT <k, m>$
\ENDMAP
\STATE $return(H)$
\end{algorithmic}
\end{algorithm}

As opposed to the dense version, the matrix generation process has be adapted to avoid visiting all possible combinations in the sparse features, using as much as possible accumulators to compute those combinations formed by zeros. As accumulators, the algorithm calculates the class histogram for the conditional variable and the joint class histogram for the parametric variables: $jind$ and $yind$. Firstly, a loop is started for those combinations in which the first variable ($i$) is not equal to zero. The procedure is the same as in the dense version. Here, if the second variable ($j$) is equal to zero, the frequency counter (in the joint histogram) for the combination will be decreased by one. Secondly, for those cases in which $j$ is not equal to zero, the algorithm completes the matrix with the frequencies in the joint histogram. And finally, for those cases in which both variables ($j$ and $y$) are equal to zero, the matrix is updated with the remaining occurrences for all classes\footnote{Class vector is always dense}. The final result will be the feature index and the aforementioned matrix.

\subsubsection{Complexity of the algorithms}
\label{subsubsec:complex}

As we mentioned before, the FS algorithm performs a greedy search which stops when the condition defined as input is reached. Beyond that, this sequential algorithm is influenced by the set of distributed algorithms/operations presented in the previous section. The distributed primitives used in these algorithms need to be analyzed to check the complexity of the whole proposal. Note that the first operation (columnar transformation) is quite time-consuming as it makes a strong use of network and memory when shuffling all data (wide dependency). However, once data have a known partitioning, they can be re-used in the following phases (leveraging from the data locality property). Anyway, this transformation is performed once at the start, and can be omitted if data are already in a columnar format. The list of distributed operations involved in this algorithm are described below:

\begin{itemize}

	\item Algorithm~\ref{alg:columnar}: This algorithm starts with a mapPartitions operation that transposes the local matrix contained in the partition, and emits a tuple for each feature (linear distributed order). The total number of tuples is equal to multiplying $n$ by the number of original partitions. Afterwards, these tuples are shuffled across the cluster, then a local sorting is launched on each subset (log-linear distributed order).
	\item Algorithm~\ref{alg:relevances}-\ref{alg:redundancies}: the first operation for both algorithms is the retrieval of a single column (feature) by using the~\emph{lookup} primitive (linear distributed order). As the data are already partitioned, the operation is done efficiently by only looking at the right partition. This variable is then broadcasted to all the nodes, which implies sending a single feature ($m$ values) across the network. The next operations (histograms and MI computations) are described below.
	\item Algorithm~\ref{alg:hist}: this algorithm represents a simple mapReduce operation where the previously generated tuples are transformed to local histograms, and finally reduced to the final histograms by feature. This map operation consists of a linear function ($O(m)$) that fetches the data contained in each local matrix.
	\item Algorithm~\ref{alg:minfo}: this operation starts by broadcasting three single values (proportion values). For each feature, three linear operations are launched to compute some extra probabilities. Finally, the mutual information values are computed by fetching the whole 3D-histogram (cubic linear order). Notice that the complexity of all these operations is bounded by the cardinality of the features included in the histogram.

\end{itemize}

\section{Experimental Framework and Analysis}
\label{sec:experiments}

This section describes the experiments carried out to evaluate the usefulness of FS over a set of real-world huge problems --both in features and examples-- using the proposed framework. 


\subsection{Datasets and methods}
\label{subsec:methods}

Five classification datasets were used in order to measure the quality and usefulness of our framework implementation for Spark. We have classified these datasets into two groups: dense (large number of samples) and sparse (high-dimensional datasets). For sparse datasets, the high-dimensional version presented in Section~\ref{subsubsec:sparse} is used. 

The first dataset~\textit{ECBDL14} was used as a reference at 
the International Conference GECCO-2014. This consists of 631 characteristics (including both numerical and categorical attributes) and 32 million instances. It is a binary classification problem where the class distribution is imbalanced: 98\% of negative instances. For this imbalanced problem, the MapReduce version of the Random OverSampling (ROS) algorithm presented in~\cite{Rio14} was applied (henceforth we will use \textit{ECBDL14} to refer to the ROS version). Another dataset used is~\textit{dna}, which consists of 50 000 000 instances with 201 discrete features. This dataset was created for the Pascal Large Scale Learning Challenge\footnote{\url{http://largescale.ml.tu-berlin.de/summary/}} in 2008. In the experiments, it was only used the training set since the test set of~\textit{dna} does not contain the class labels, the training set was used to generate both subsets (using an 80/20 hold-out data split). As this problem also suffers imbalance between its classes, ROS technique was also applied (henceforth \textit{dna}). The rest of datasets (epsilon, url and kddb) come from the LibSVM dataset repository~\cite{chang11}. These datasets and their descriptions can be found in the project's website\footnote{\url{http://www.csie.ntu.edu.tw/~cjlin/libsvmtools/datasets/}}. Table~\ref{tab:datasets} gives a summary of these datasets. 

\begin{table}[!htp]
\renewcommand{\arraystretch}{1.3}
\centering
\scriptsize
\caption{Summary description of employed datasets. For each one, the number of examples for train and test sets (\#Train Ex., \#Test Ex.), the total number of attributes (\#Atts.), the number classes (\#Cl) and its sparsity condition (Sparse) are shown.}
\label{tab:datasets}
\resizebox{0.75\textwidth}{!}{
  \begin{tabular}{|l||c||c||c||c||c|}
  \hline {\bf Data Set} & {\bf \#Train Ex.} & {\bf \#Test Ex.} & {\bf \#Atts.}  &{\bf \#Cl.} & {\bf Sparse}\\
  \hline
epsilon & 400 000 & 100 000 & 2000 & 2 & No \\
dna & 79 739 293 & 10 000 000 & 200 & 2 & No\\
ECBDL14 & 65 003 913 & 2 897 917 & 630 & 2 & No\\
url & 1 916 904 & 479 226 & 3 231 961 & 2 & Yes\\
kddb & 19 264 097 & 748 401 & 29 890 095 & 2 & Yes\\
  \hline
  \end{tabular}
}
\end{table}

As an FS benchmark method, we have used mRMR algorithm~\cite{peng2005} since it is one of the most relevant and cited selectors in the literature. Note that the FS algorithm chosen to test the performance does not affect the time results yielded by the framework since all criteria are computed in the same way. 

In order to carry out a comparison study, the following classifiers were chosen: Support Vector Machines (SVM)~\cite{Hearst1998}, and Naive Bayes~\cite{Duda1973}. For the experiments, we have employed the distributed versions of these algorithms implemented in the MLlib library~\cite{mll15}. The recommended parameters of the classifiers, according to their authors' specification~\cite{mll15}, are shown in Table~\ref{tab:parameters}. For all executions, the datasets have been cached in memory as SVM and our method are iterative processes.
The level of parallelism (number of partitions) has been set to 864, twice the total number of cores available in the cluster\footnote{The Spark creators recommend using 2-4 partitions per core:~\url{http://spark.apache.org/docs/latest/programming-guide.html}}.

\begin{table}[!htp]
\renewcommand{\arraystretch}{1.3}
\centering
\scriptsize
\caption{Parameters of the used classifiers}
\label{tab:parameters}
\resizebox{0.9\textwidth}{!}{
  \begin{tabular}{|l|c|}
  \hline
  \textbf{Method} & \textbf{Parameters}\\
  \hline
  Naive Bayes & lambda = 1.0\\
  SVM & stepSize = 1.0, batchFraction = 1.0, regularization = 1.0, iterations = 100 \\
  \hline
  mRMR & level of parallelism = 864\\
  \hline
  \end{tabular}
  }
\end{table}

A Spark package associated to this work can be found in the third-party Spark's Repository: \textit{\url{http://spark-packages.org/package/sramirez/spark-infotheoretic-feature-selection}}. This software has been designed to be integrated as part of MLlib Library. This has associated a JIRA issue to discuss its integration in this library: \textit{\url{https://issues.apache.org/jira/browse/SPARK-6531}}.

For evaluation purposes, we use two common evaluation metrics to assess the quality of the subsequent FS schemes: Area under Receiver Operating Characteristic (AUROC, henceforth called only AUC) to evaluate the accuracy yielded by the classifier, and the modeling time in training to evaluate the performance of the FS process.

%
%
%
%

\subsection{Cluster configuration}
\label{subsec:cluster}

For all the experiments we have used a cluster composed of eighteen computing nodes and one master node. The computing nodes hold the following characteristics: 2 processors x Intel Xeon CPU E5-2620, 6 cores per processor, 2.00 GHz, 15 MB cache, QDR InfiniBand Network (40 Gbps), 2 TB HDD, 64 GB RAM. Regarding the software, we have used the following configuration: Hadoop 2.5.0-cdh5.3.1 from Cloudera's open-source Apache Hadoop distribution\footnote{http://www.cloudera.com/content/cloudera/en/documentation/cdh5/v5-0-0/CDH5-homepage.html}, HDFS replication factor: 2, HDFS default block size: 128 MB, Apache Spark and MLlib 1.2.0, 432 cores (24 cores/node), 864 RAM GB (48 GB/node).

%

Both HDFS and Spark master processes (the HDFS NameNode and the Spark Master) are hosted in the main node. The NameNode controls the HDFS, coordinating the slave machines by the means of their respective DataNode daemons whereas the Spark Master controls all the executors in each worker node. Spark uses HDFS file system to load and save data in the same way Hadoop framework does. 


\subsection{Analysis of selection results}
\label{subsec:results-time}

Here, we evaluate the time employed by our implementation to rank the most relevant features. Table~\ref{tab:fs-time} presents the time results obtained by our algorithm using different ranking thresholds (number of features selected).

\begin{table}[!htp]
\renewcommand{\arraystretch}{1.3}
\centering
\scriptsize
\caption{Selection time by dataset and threshold (in seconds)}
\label{tab:fs-time}
\resizebox{0.9\textwidth}{!}{
\begin{tabular}{cccccc}
\hline
 \textbf{\# Features} & \textbf{kddb} & \textbf{url} & \textbf{dna} & \textbf{ECBDL14} & \textbf{epsilon}\\ 
\hline
10 & 283.61 & 94.06 & 97.83 & 332.90 & 111.42\\ 
25 & 774.43 & 186.22 & 148.78 & 596.31 & 173.39\\ 
50 & 1365.82 & 333.70 & 411.84 & 1084.58 & 292.07\\ 
100 & 2789.55 & 660.48 & 828.35 & 2420.94 & 542.05\\ 
\hline
\end{tabular}
}
\end{table}


As can be seen in Table~\ref{tab:fs-time}, our algorithm yields competitive results in all cases regardless of the number of iterations employed (represented by the threshold value). For those datasets with the highest volume of data: kddb (ultra-high dimensional data) and ECBDL14 (huge number of samples), it is important to note that our method is able to rank 100 features in less than one hour. 

Furthermore, a comparison study is performed between our distributed version and the sequential version developed by Brown's lab\footnote{FEAST toolbox (python version):~\url{http://www.cs.man.ac.uk/~gbrown/software/}}. Samples from~\textit{dna} have been generated with different ratios of instances to study the scalability of our approach in counterpart to the sequential version\footnote{Sequential version has been executed in one node of our cluster with the aforementioned characteristics}. The level of parallelism has been set to 200 in the distributed executions. This has been done to ease the comparison between the distributed version which uses one core per feature, and the sequential version that only uses one core. 

\begin{figure}[!thp]
  \begin{center}
    \includegraphics[width=0.75\textwidth]{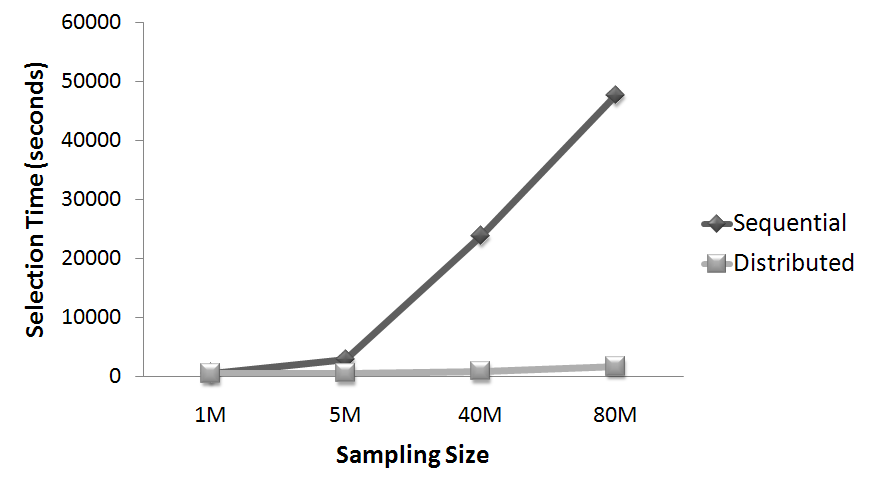}
  \caption{Selection time comparison (sequential vs. distributed)}
  \label{fig:fs-comp}
  \end{center}
\end{figure}

Table~\ref{tab:fs-comp} and Figure~\ref{fig:fs-comp} show the time results for this comparison. Regarding the sequential version, the last two values was estimated by using linear interpolation (highlighted in italics) since they could not be computed due to memory problems. As shown in Table~\ref{tab:fs-comp}, our distributed version outperforms the classical approach in all cases. This is specially remarkable for the largest dataset where the maximum speedup rate is achieved (29.83).

\begin{table}[!htp]
\renewcommand{\arraystretch}{1.3}
\centering
\scriptsize
\caption{Selection time comparison (sequential vs. distributed) in seconds.}
\label{tab:fs-comp}
\resizebox{0.9\textwidth}{!}{
\begin{tabular}{lccc}
\hline
 \textbf{\# Examples} & \textbf{Sequential} & \textbf{Distributed} & \textbf{Speedup}\\ 
\hline
1 000 000 & 450.00 & 425.56 & 1.06 \\ 
5 000 000 & 2 839.72 & 508.41 & 5.59 \\
40 000 000 & \textit{23 749.77} & 828.35 & 28.67\\
80 000 000 & \textit{47 646.97} & 1 597.26 & \textbf{29.83}\\
\hline
\end{tabular}
}
\end{table}

\begin{figure}[!thp]
  \begin{center}
    \includegraphics[width=0.75\textwidth]{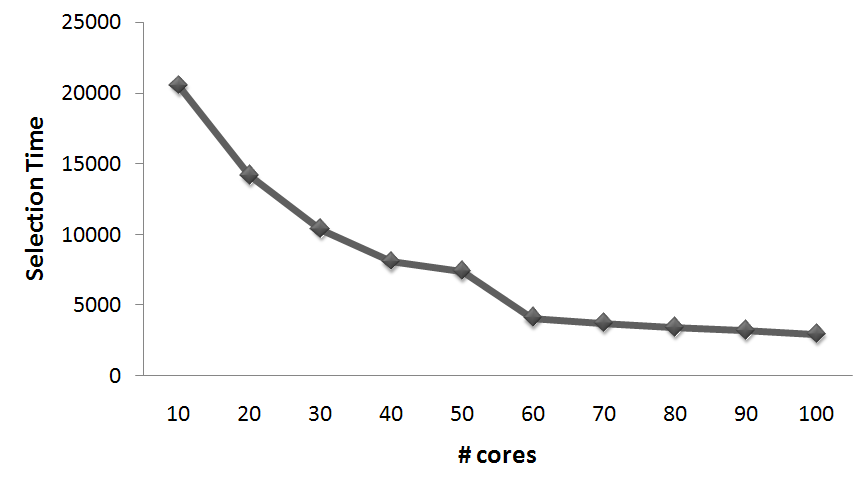}
  \caption{Selection time by number of cores (in seconds).}
  \label{fig:fs-speedup}
  \end{center}
\end{figure}

Finally, an additionaly study of scalability has been performed by varying the number of cores used. For this study,~\textit{ECBDL14} (the largest dense dataset) was used as a reference, with the same parameters as in the previous study. Figure~\ref{fig:fs-speedup} depicts the performance of our method varying the number of cores from ten to one hundred. The results show a logarithmic behaviour as the number of cores is increased. Note that in the first case, the number of nodes is only ten and the amount of memory available is not the fullest.

\subsection{Analysis of classification results}
\label{subsec:results-cls}

\begin{figure}[!thp]
  \begin{center}
    \includegraphics[width=0.75\textwidth]{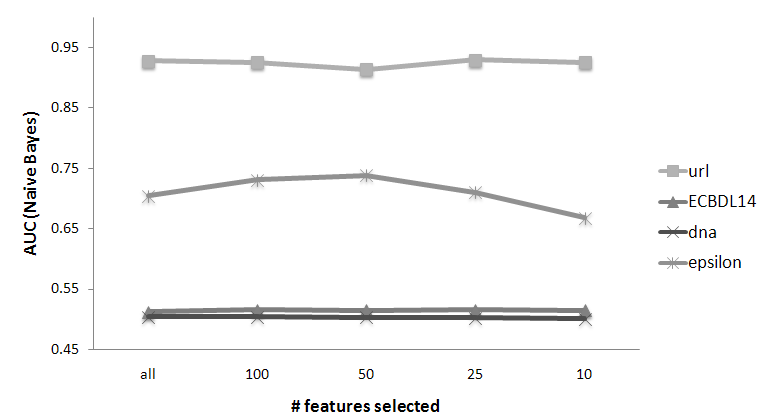}
  \caption{AUC results for NaiveBayes using different thresholds}
  \label{fig:fs-nb}
  \end{center}
\end{figure}

\begin{figure}[!thp]
  \begin{center}
    \includegraphics[width=0.75\textwidth]{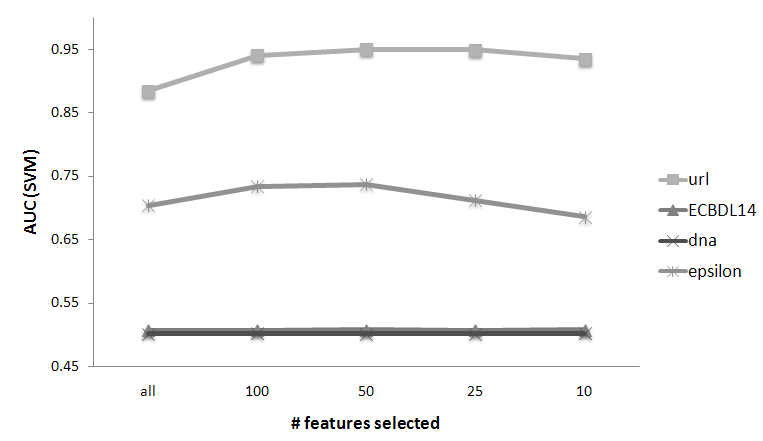}
  \caption{AUC results for SVM using different thresholds}
  \label{fig:fs-svm}
  \end{center}
\end{figure}

In this section, a study about the usefulness of our FS solution on large-scale classification is performed. Figures~\ref{fig:fs-nb} and~\ref{fig:fs-svm} show the accuracy results for SVM and Naive Bayes using different FS schemes. All datasets described in~Table~\ref{tab:datasets} have been used in this study except~\textit{kddb} because the aforementioned classifiers are not designed to work with such a big dimensionality. 

Figure~\ref{fig:fs-svm} shows an important improvement on using FS over~\textit{url} and~\textit{epsilon}, whereas its application seems to have a negligible impact on AUC for~\textit{dna} and~\textit{ECBDL14}. This can be explained by the fact of their high imbalance ratio and/or their low number of features. Figure~\ref{fig:fs-nb} presents similar results to the previous case. However, in this case the improvement in~\textit{url} dataset is much smaller.

\begin{figure}[!thp]
  \begin{center}
    \includegraphics[width=0.75\textwidth]{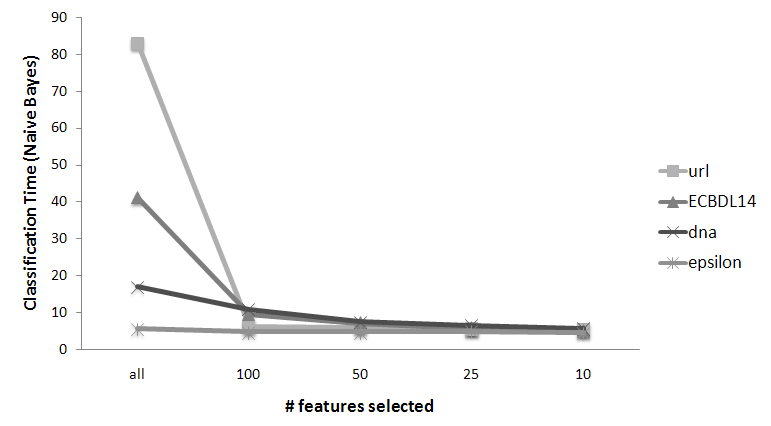}
  \caption{Classification time results in training for NaiveBayes using different thresholds (in seconds)}
  \label{fig:fs-nb-time}
  \end{center}
\end{figure}

\begin{figure}[!thp]
  \begin{center}
    \includegraphics[width=0.75\textwidth]{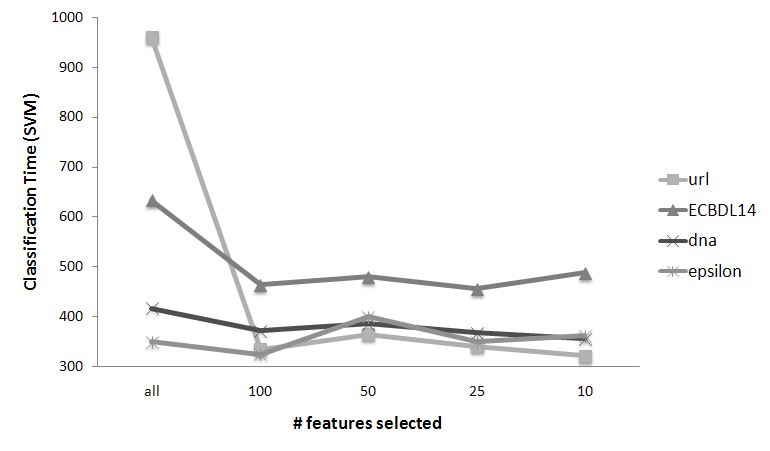}
  \caption{Classification time results in training for SVM using different thresholds (in seconds)}
  \label{fig:fs-svm-time}
  \end{center}
\end{figure}

Beyond AUC, the time employed in creating a classification model is quite important in many large-scale problems. Figures~\ref{fig:fs-nb-time} and~\ref{fig:fs-svm-time} show the classification time employed in the training phase for different datasets and thresholds. The results demonstrate that the simplicity and the performance of the generated models is improved after applying FS in all the cases studied. It is specially important for SVM, which spend more time modeling than Naive Bayes.

\bigskip

Results in performance demonstrate that our solution is capable of selecting features in a competitive time when it is applied to huge datasets --both in number of instances and features--. Results also demonstrate the benefit of using our approach against the sequential version in all cases, enabling the application of FS on the largest cases.

Furthermore, using our selection schemes the classifiers show better classification results in most cases, obtaining similar results in the other cases. Note that in all the studied cases the resulting model is much simpler and faster in spite of using a small percentage of the original set of features.

\section{Conclusions}
In this paper, we have discussed the problem of processing huge data, especially from the perspective of dimensionality. We have seen the effects of a correct identification of relevant features on these datasets as well as the difficulty of this task due to the combinatorial effects when incoming data grow --both in number instances and features--. In spite of the growing interest in the field of dimensionality reduction for Big Data, only few FS methods have been developed to deal with high-dimensional problems.

Thereby we have redesigned a generic FS framework for Big Data based on Information Theory, adapting a previous one proposed by Brown \textit{et al.} in~\cite{brown12}. The framework contains implementations of many state-of-the-art FS algorithms like mRMR or JMI. However, the adaptation carried out has entailed a deep redesign of Brown \textit{et al.}'s framework so as to adapt it to the distributed paradigm. With this work we have also aimed at contributing by adding an FS module to the emerging Spark and MLlib platforms, where no complex FS algorithm has been included until now.

The experimental results show the usefulness of our FS solution over a wide set of large real-world problems. Our solution has thus revealed to perform well with two dimensions of Big Data (samples and features), obtaining competitive performance results when dealing with ultra-high dimensional datasets as well as those with a huge number of samples. Furthermore, our solution has outperformed the sequential version in all studied cases, enabling the resolution of problems that were not practical with the classical approach.

\section*{Acknowledgment}
This work is supported by the Spanish National Research Project TIN2012-37954, TIN2013-47210-P and TIN2014-57251-P, and the Andalusian Research Plan P10-TIC-6858, P11-TIC-7765 and P12-TIC-2958, and by the Xunta de Galicia through the research project GRC 2014/035 (all projects partially funded by FEDER funds of the European Union). S. Ram\'irez-Gallego holds a FPU scholarship from the Spanish Ministry of Education and Science (FPU13/00047). D. Mart\'inez-Rego and V. Bol\'on-Canedo acknowledge support of the Xunta de Galicia under postdoctoral Grant codes POS-A/2013/196 and ED481B 2014/164-0.





%
\bibliographystyle{plain}
\bibliography{biblio}

\begin{thebibliography}{10}
\bibitem{lin12}
Jimmy Lin.
\newblock Mapreduce is good enough? if all you have is a hammer, throw away
  everything that's not a nail!
\newblock {\em CoRR}, abs/1209.2191, 2012.
\end{thebibliography}

%

%
%
%




\end{document}